\pdfoutput=1
\documentclass[twoside,11pt]{article}

\usepackage[abbrvbib, preprint]{jmlr2e}

\usepackage{listings}
\usepackage{multirow}
\usepackage{algorithmic}
\usepackage{algorithm}
\usepackage{booktabs}
\usepackage{hyperref}
\usepackage{xspace}
\usepackage{caption}
\usepackage{subcaption}

\usepackage{color}
\usepackage{xcolor}

\usepackage{tikz}

\newcommand{\awnas}{aw\_nas\xspace}

\ShortHeadings{aw\_nas: A Modularized and Extensible NAS framework}{Ning, Tang, Li, Yang, Zhao, Zhang, Lu, Liang, Yang, Wang}
\firstpageno{1}

\begin{document}

\title{aw\_nas: A Modularized and Extensible NAS framework}

\author{\name Xuefei Ning$^1$ \email foxdoraame@gmail.com \\
  \name Changcheng Tang$^2$ \email changcheng.tang@novauto.com.cn \\
  \name Wenshuo Li$^1$ \email wilsonleethu@gmail.com \\
  \name Songyi Yang$^2$ \email patrick22414@outlook.com \\
  \name Tianchen Zhao$^2$ \email ztc16@buaa.edu.cn\\
  \name Niansong Zhang$^2$ \email 
  nz264@cornell.edu\\
  \name Tianyi Lu$^2$ \email tianyi.lu@novauto.com.cn \\
  \name Shuang Liang$^{12}$ \email shuang.liang@novauto.com.cn \\
  \name Huazhong Yang$^{1}$\thanks{Corresponding authors.} \email yanghz@mail.tsinghua.edu.cn \\
  \name Yu Wang$^{1*}$ \email yu-wang@tsinghua.edu.cn \\
  \addr $^1$ Department of Electronic Engineering, Tsinghua University, Beijing, China\\
  \addr $^2$ Novauto Co. Ltd., Beijing, China}

\maketitle

\begin{abstract}

  
  Neural Architecture Search (NAS) has received extensive attention due to its capability to discover neural network architectures in an automated manner.
  aw\_nas is an open-source Python framework implementing various NAS algorithms in a modularized manner.
  Currently, \awnas can be used to reproduce the results of mainstream NAS algorithms of various types. 
  Also, due to the modularized design, one can simply experiment with different NAS algorithms for various applications with \awnas (e.g., classification, detection, text modeling, fault tolerance, adversarial robustness, hardware efficiency, and etc.).
Codes and documentation are available at \url{
https://github.com/walkerning/aw\_nas}.

\end{abstract}


\section{Introduction}

Neural Architecture Search (NAS) has received extensive attention due to its 
capability to discover competitive neural network architectures in an automated manner.
Early NAS algorithms~\citep{nasnet,real2019regularized} are extremely slow, since a separate training phase is needed to evaluate each architecture, and tons of candidate architectures need to be evaluated to explore the large search space.
Major efforts to alleviate the computational challenge of NAS lie in three aspects: 1) Better and compact search space design~\citep{zoph2018learningta}. 2) Accelerate the evaluation of each candidate architecture~\citep{baker2017accelerating,elsken2018efficient,enas}; 3) Improve the sample efficiency of search space exploration~\citep{kandasamy2018bayesian,ning2020generic}.

Those methods that aim to accelerate architecture evaluation can be further categorized according to whether or not the separate training phase is still needed for each architecture. Early studies shorten the separate training phase of each architecture by training curve extrapolation~\citep{baker2017accelerating}, good weight initialization~\citep{elsken2018efficient}, and so on.
On the other hand, the current trending practice, parameter-sharing evaluation, is to amortize architectures' training to the training of a shared set of parameters~\citep{darts,enas,cai2020once}, thus avoid separately training each candidate architecture. From the aspect of improving the sample efficiency, a promising direction is to use predictor-based NAS methods~\citep{kandasamy2018bayesian,ning2020generic}. 
These methods learn a performance predictor and utilize its predictions to select architectures that are more worth evaluating.



\section{\awnas Description}
\awnas aims to provide a general, extensible and easy-to-use NAS framework, so that not only researchers can build and compare their methods in a more controlled setting, but nonprofessionals can also easily apply NAS techniques to their specific applications.

\subsection{Framework Design}
\label{sec:framework}

The main design principle lying behind \awnas is modularization.
There are multiple actors that are working together in a NAS algorithm, and they can be categorized into well-defined components based on their roles.
The list of components and the \awnas supported choices for each component are summarized in Tab.~\ref{table:components}.

\begin{table}[ht]
    \vspace{-5pt}
  \centering
  \caption{\awnas supported component types}
  \label{table:components}
  \begin{tabular}{c|p{4cm}p{7cm}}
    \toprule
    {Component}  & {Description} & {Current supported types} \\
    \midrule
    \multirow{2}{*}{Dataset} & define the dataset & Cifar-10/100, SVHN, (Tiny-)ImageNet, PTB, VOC, COCO, TT100k\\
    \hline
    \multirow{4}{*}{Objective} & the rewards to learn the controller, and (optionally) the objectives to update the evaluator &  classification, detection, language, fault tolerance, adversarial robustness, hardware (latency, energy ...)\\
    \hline
    \multirow{4}{*}{Search space} & define what architectural decision to be made & cell-based CNN, dense cell-based CNN, cell-based RNN, NasBench-101/201, blockwise with mnasnet/mobilenet backbones\\
    \hline
    \multirow{3}{*}{Controller}  & select architectures to be evaluated & random sample, simulated annealing, evolutionary, RL-learned sampler, differentiable, predictor-based\\
    \hline
    \multirow{2}{*}{Weights manager} & fill the architectures with weights & supernet, differentiable supernet, morphism-based\\
    \hline
    \multirow{2}{*}{Evaluator} & how to evaluate an architecture & parameter-sharing evaluator (mepa), separately tune and evaluate (tune)\\
    \hline
    \multirow{3}{*}{Trainer} & the orchestration of the overall NAS search flow & a general workflow described in Sec.\ref{sec:framework} (simple), parallelized evaluation and async update of controller (async)\\
    \bottomrule
  \end{tabular}
\end{table}

The interface between these components is well-defined. We use a ``rollout'' (class \textit{awnas.rollout.base.BaseRollout}) to represent the interface object between all these components. Usually, a search space defines one or more rollout types (a subclass of \textit{BaseRollout}). For example, the basic cell-based search space \textbf{cnn}
corresponds to two rollout types: 1) \textbf{discrete} rollouts that are used in reinforcement learning (RL) based, evolutionary based controllers, and etc. 
2) \textbf{differentiable} rollouts that are used in gradient-based NAS. 

\begin{figure}[tb]
\begin{center}
  \includegraphics[width=0.9\linewidth]{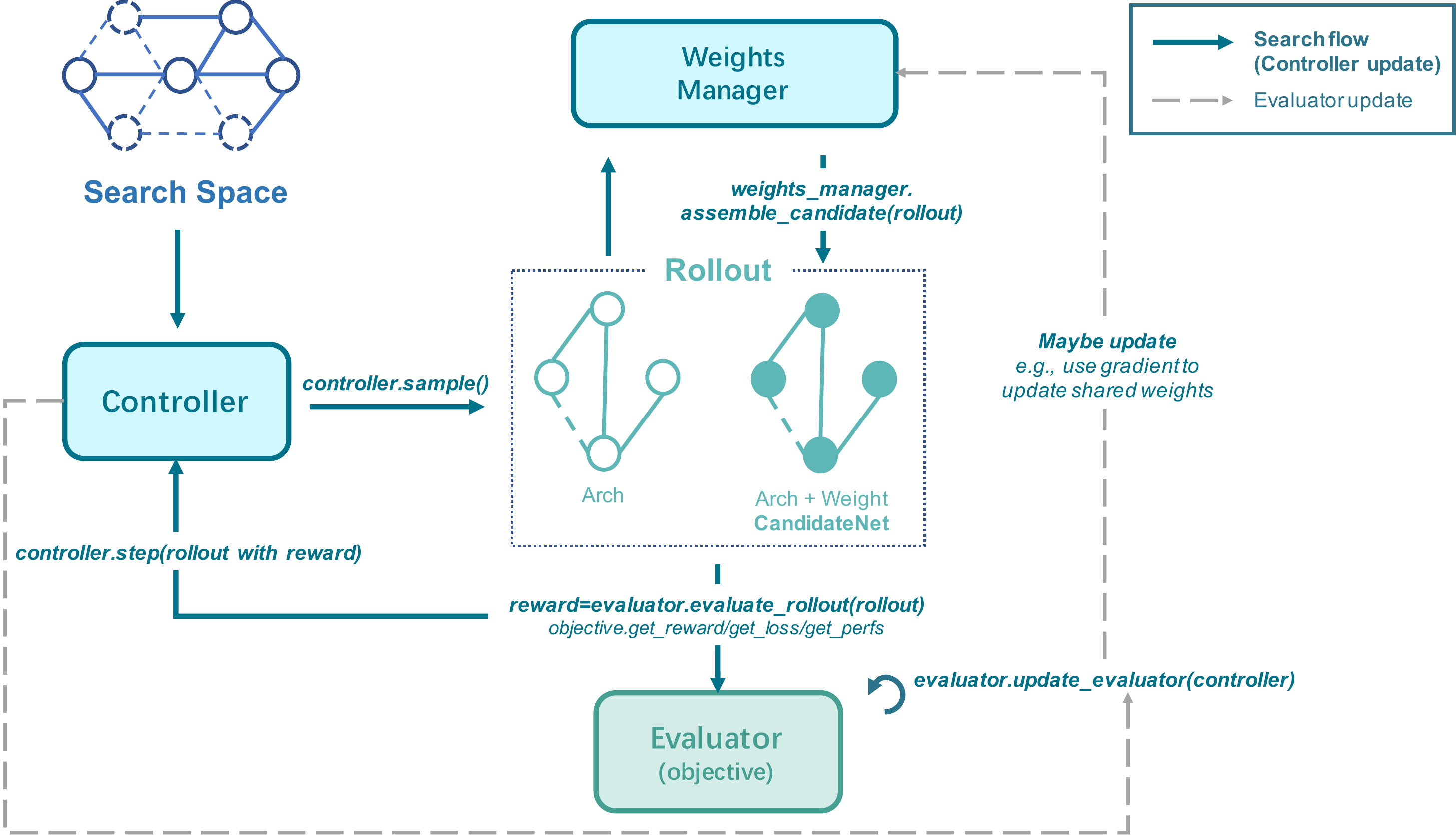}
  \caption{Search workflow and interfaces.}
\label{fig:workflow}
\end{center}
\vspace{-5pt}
\end{figure}

The search workflow of a NAS algorithm and some important interface methods are illustrated in Fig.~\ref{fig:workflow}. Specifically, one iteration of the search flow goes as follows:
\begin{enumerate}
\item \textit{rollout = controller.sample()}: The \texttt{controller} is responsible for sampling candidate architectures from the search space.
\item \textit{weights\_manager.assemble\_candidate(rollout)}: The weights manager fills the sampled architecture with weights.
\item \textit{evaluator.evaluate\_rollout(rollout)}: The evaluator evaluate the rollout that contains the architecture and weights information.
\item \textit{controler.step(rollout)}: The rollout that contains the reward information is used to update the controller.
\item Optionally, some types of evaluator might need to be updated periodically by calling \textit{evaluator.update\_evaluator(controller)}, which might issue calls to \textit{controller.sample} \textit{weights\_manager.assemble\_candidate} too.
\end{enumerate}

Taking the ENAS~\citep{enas} method as an example, the \texttt{dataset} and \texttt{objective} are of type \textbf{cifar10} and \textbf{classification}, respectively. The \texttt{search space} type \textbf{cnn} defines a cell-based CNN search space. And the \texttt{controller} \textbf{rl} is a RL-learned RNN network. The \texttt{weights\_manager} \textbf{supernet} is a parameter-sharing based supernet. As for the \texttt{evaluator} \textbf{mepa}, with its most basic configuration, just forward batches sampled from the dataset and call \textit{objective.get\_reward} to get the rollout's reward.



\begin{table}[tb]
  \vspace{-10pt}
  \centering
  \caption{\awnas command-line utilities}
  \label{tab:command}
  \begin{tabular}{c|c}
    \toprule
    {Subcommand} & {Description} \\
    \midrule
    search / mpsearch      &  (Multiprocessing) Search for architecture\\
    random-sample          &  Random sample architectures\\
    sample                 &  Sample architectures with a controller\\
    derive                 &  Derive architectures with trained NAS components\\
    eval-arch              &  Eval architectures in a YAML file with an evaluator\\
    train / mptrain / test      &  (Multiprocessing) Train or test an architecture\\
    gen-(final-)sample-config      &  Dump the sample configuration for search (final training) \\
    registry               &  Print registry information\\
    \bottomrule
  \end{tabular}
  \vspace{-10pt}
\end{table}

\subsection{Basic Usage}

\awnas standardize a typical NAS workflow into a 3-step process, i.e., \textbf{search-derive-train}.
After the search phase that is described in Sec.~\ref{sec:framework},
the derive utility makes architecture decision using the trained NAS components. 
Then, a final training phase is conducted to train and evaluate the derived architecture.
With \awnas, combining various components and run a NAS algorithm is no more than just tweaking several configuration files and then run the command-line tool \textit{awnas} with it. Currently, in \awnas version 0.4, the available subcommands of the \textit{awnas} command-line tool are summarized in Tab.~\ref{tab:command}.

  




\section{Conclusion and Future Work}

We introduce \awnas, a modularized and extensible framework for NAS algorithms. By implementing various types of NAS components in a modularized way, \awnas allows users to pick up components and run a NAS algorithm easily. An unified implementation with clear interface design also makes it easier for researchers and developers to develop and compare new NAS methods. 

\awnas is still under active development. We are trying to scale it to applications with larger scales, and make it easier for nonprofessionals to build up effective NAS systems targeted for their specific application scenarios.

\acks{This work was supported by National Key Research and Development Program of China (No. 2018YFB0105000),
  National Natural Science Foundation of China (No. U19B2019, 61832007), Beijing National Research Center for Information Science and Technology (BNRist).
This framework also contains contributions made by students in NICS-EFC laboratory of Tsinghua University: Zixuan Zhou, Junbo Zhao, Shulin Zeng.}


\newpage

\vskip 0.2in
\bibliographystyle{splncs}
\bibliography{sample}

\begin{thebibliography}{21}
\providecommand{\natexlab}[1]{#1}
\providecommand{\url}[1]{\texttt{#1}}
\expandafter\ifx\csname urlstyle\endcsname\relax
  \providecommand{\doi}[1]{doi: #1}\else
  \providecommand{\doi}{doi: \begingroup \urlstyle{rm}\Url}\fi

\bibitem[Baker et~al.(2017)Baker, Gupta, Raskar, and
  Naik]{baker2017accelerating}
B.~Baker, O.~Gupta, R.~Raskar, and N.~Naik.
\newblock Accelerating neural architecture search using performance prediction.
\newblock \emph{arXiv preprint arXiv:1705.10823}, 2017.

\bibitem[Cai et~al.(2020)Cai, Gan, Wang, Zhang, and Han]{cai2020once}
H.~Cai, C.~Gan, T.~Wang, Z.~Zhang, and S.~Han.
\newblock Once for all: Train one network and specialize it for efficient
  deployment.
\newblock In \emph{International Conference on Learning Representations}, 2020.

\bibitem[Chen et~al.(2019)Chen, Yang, Zhang, Meng, Xiao, and Sun]{detnas}
Y.~Chen, T.~Yang, X.~Zhang, G.~Meng, X.~Xiao, and J.~Sun.
\newblock Detnas: Backbone search for object detection.
\newblock In \emph{NeurIPS}, 2019.

\bibitem[Elsken et~al.(2018)Elsken, Metzen, and Hutter]{elsken2018efficient}
T.~Elsken, J.~H. Metzen, and F.~Hutter.
\newblock Efficient multi-objective neural architecture search via lamarckian
  evolution.
\newblock In \emph{International Conference on Learning Representations}, 2018.

\bibitem[Everingham et~al.(2009)Everingham, Gool, Williams, Winn, and
  Zisserman]{voc}
M.~Everingham, L.~Gool, C.~K. Williams, J.~Winn, and A.~Zisserman.
\newblock The pascal visual object classes (voc) challenge.
\newblock \emph{International Journal of Computer Vision}, 88:\penalty0
  303--338, 2009.

\bibitem[Howard et~al.(2019)Howard, Sandler, Chu, Chen, Chen, Tan, Wang, Zhu,
  Pang, Vasudevan, et~al.]{mobilenetv3}
A.~Howard, M.~Sandler, G.~Chu, L.-C. Chen, B.~Chen, M.~Tan, W.~Wang, Y.~Zhu,
  R.~Pang, V.~Vasudevan, et~al.
\newblock Searching for mobilenetv3.
\newblock In \emph{Proceedings of the IEEE International Conference on Computer
  Vision}, pages 1314--1324, 2019.

\bibitem[Kandasamy et~al.(2018)Kandasamy, Neiswanger, Schneider, Poczos, and
  Xing]{kandasamy2018bayesian}
K.~Kandasamy, W.~Neiswanger, J.~Schneider, B.~Poczos, and E.~P. Xing.
\newblock Neural architecture search with bayesian optimisation and optimal
  transport.
\newblock In \emph{Advances in Neural Information Processing Systems}, pages
  2016--2025, 2018.

\bibitem[Li et~al.(2020)Li, Ning, Ge, Chen, Wang, and Yang]{li2020fttnas}
W.~Li, X.~Ning, G.~Ge, X.~Chen, Y.~Wang, and H.~Yang.
\newblock Ftt-nas: Discovering fault-tolerant neural architecture.
\newblock In \emph{2020 25th Asia and South Pacific Design Automation
  Conference (ASP-DAC)}, pages 211--216, 2020.

\bibitem[Liu et~al.(2018)Liu, Simonyan, and Yang]{darts}
H.~Liu, K.~Simonyan, and Y.~Yang.
\newblock Darts: Differentiable architecture search.
\newblock \emph{arXiv preprint arXiv:1806.09055}, 2018.

\bibitem[Liu et~al.(2016)Liu, Anguelov, Erhan, Szegedy, Reed, Fu, and
  Berg]{liu2016ssd}
W.~Liu, D.~Anguelov, D.~Erhan, C.~Szegedy, S.~Reed, C.-Y. Fu, and A.~C. Berg.
\newblock Ssd: Single shot multibox detector.
\newblock In \emph{European conference on computer vision}, pages 21--37.
  Springer, 2016.

\bibitem[Ning et~al.(2020{\natexlab{a}})Ning, Li, Zhou, Zhao, Zheng, Liang,
  Yang, and Wang]{ning2020surgery}
X.~Ning, W.~Li, Z.~Zhou, T.~Zhao, Y.~Zheng, S.~Liang, H.~Yang, and Y.~Wang.
\newblock A surgery of the neural architecture evaluators, 2020{\natexlab{a}}.

\bibitem[Ning et~al.(2020{\natexlab{b}})Ning, Zheng, Zhao, Wang, and
  Yang]{ning2020generic}
X.~Ning, Y.~Zheng, T.~Zhao, Y.~Wang, and H.~Yang.
\newblock A generic graph-based neural architecture encoding scheme for
  predictor-based nas.
\newblock In \emph{European Conference on Computer Vision}, 2020{\natexlab{b}}.

\bibitem[Pham et~al.(2018)Pham, Guan, Zoph, Le, and Dean]{enas}
H.~Pham, M.~Y. Guan, B.~Zoph, Q.~V. Le, and J.~Dean.
\newblock Efficient neural architecture search via parameter sharing.
\newblock In \emph{International Conference on Machine Learning (ICML)}, 2018.

\bibitem[Real et~al.(2019)Real, Aggarwal, Huang, and Le]{real2019regularized}
E.~Real, A.~Aggarwal, Y.~Huang, and Q.~V. Le.
\newblock Regularized evolution for image classifier architecture search.
\newblock In \emph{Proceedings of the aaai conference on artificial
  intelligence}, volume~33, pages 4780--4789, 2019.

\bibitem[Tang et~al.(2019)Tang, Feng, Shao, Kuang, Zhang, and
  Chen]{tang2019learning}
S.~Tang, L.~Feng, W.~Shao, Z.~Kuang, W.~Zhang, and Y.~Chen.
\newblock Learning efficient detector with semi-supervised adaptive
  distillation.
\newblock \emph{arXiv preprint arXiv:1901.00366}, 2019.

\bibitem[Wu et~al.(2019)Wu, Dai, Zhang, Wang, Sun, Wu, Tian, Vajda, Jia, and
  Keutzer]{fbnet}
B.~Wu, X.~Dai, P.~Zhang, Y.~Wang, F.~Sun, Y.~Wu, Y.~Tian, P.~Vajda, Y.~Jia, and
  K.~Keutzer.
\newblock Fbnet: Hardware-aware efficient convnet design via differentiable
  neural architecture search.
\newblock \emph{2019 IEEE/CVF Conference on Computer Vision and Pattern
  Recognition (CVPR)}, pages 10726--10734, 2019.

\bibitem[Xie et~al.(2019)Xie, Zheng, Liu, and Lin]{snas}
S.~Xie, H.~Zheng, C.~Liu, and L.~Lin.
\newblock {SNAS}: stochastic neural architecture search.
\newblock In \emph{International Conference on Learning Representations}, 2019.
\newblock URL \url{https://openreview.net/forum?id=rylqooRqK7}.

\bibitem[Xu et~al.(2019)Xu, Xie, Zhang, Chen, Qi, Tian, and Xiong]{pcdarts}
Y.~Xu, L.~Xie, X.~Zhang, X.~Chen, G.-J. Qi, Q.~Tian, and H.~Xiong.
\newblock Pc-darts: Partial channel connections for memory-efficient
  differentiable architecture search.
\newblock \emph{ArXiv}, abs/1907.05737, 2019.

\bibitem[Zeng et~al.(2020)Zeng, Sun, Xing, Ning, Shan, Chen, Wang, and zhong
  Yang]{zeng2020blackbs}
S.~Zeng, H.~Sun, Y.~Xing, X.~Ning, Y.~Shan, X.~Chen, Y.~Wang, and H.~zhong
  Yang.
\newblock Black box search space profiling for accelerator-aware neural
  architecture search.
\newblock \emph{2020 25th Asia and South Pacific Design Automation Conference
  (ASP-DAC)}, pages 518--523, 2020.

\bibitem[Zoph and Le(2017)]{nasnet}
B.~Zoph and Q.~V. Le.
\newblock Neural architecture search with reinforcement learning.
\newblock In \emph{International Conference on Learning Representations
  (ICLR)}, 2017.

\bibitem[Zoph et~al.(2018)Zoph, Vasudevan, Shlens, and Le]{zoph2018learningta}
B.~Zoph, V.~Vasudevan, J.~Shlens, and Q.~V. Le.
\newblock Learning transferable architectures for scalable image recognition.
\newblock \emph{2018 IEEE/CVF Conference on Computer Vision and Pattern
  Recognition}, pages 8697--8710, 2018.

\end{thebibliography}

\newpage

\appendix
\section*{Appendix A. Some Reproducing Results and Our Researches}


\awnas can be used to reproduce many NAS algorithms by combining different components and tweaking the configurations, and some representative studies are ENAS~\citep{enas}, DARTS~\citep{darts}, SNAS~\citep{snas}, PC-DARTS~\citep{pcdarts}, FBNet~\citep{fbnet}, OFA~\citep{cai2020once}, GATES~\citep{ning2020generic}, DetNAS~\citep{detnas}, and other traditional NAS methods. We hope that, by providing a unified and modularized code base, NAS algorithms can be compared in a more controlled setting. As an example, Tab.~\ref{tab:reproduce} shows the reproduction results of some popular parameter-sharing NAS methods.

\begin{figure}[ht]
    \makebox[\textwidth][c]{
        \includegraphics[width=0.5\textwidth]{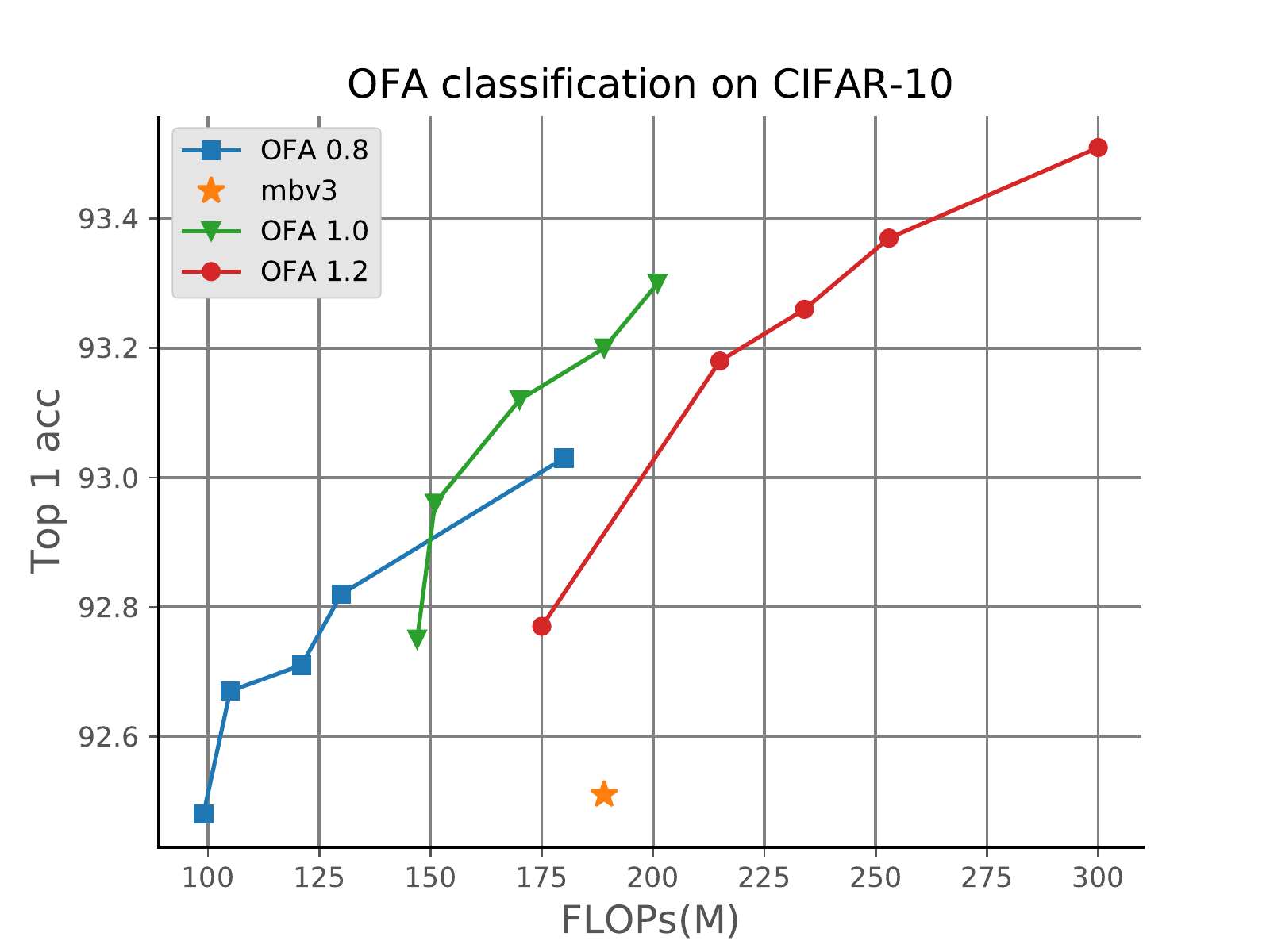}
        \includegraphics[width=0.5\textwidth]{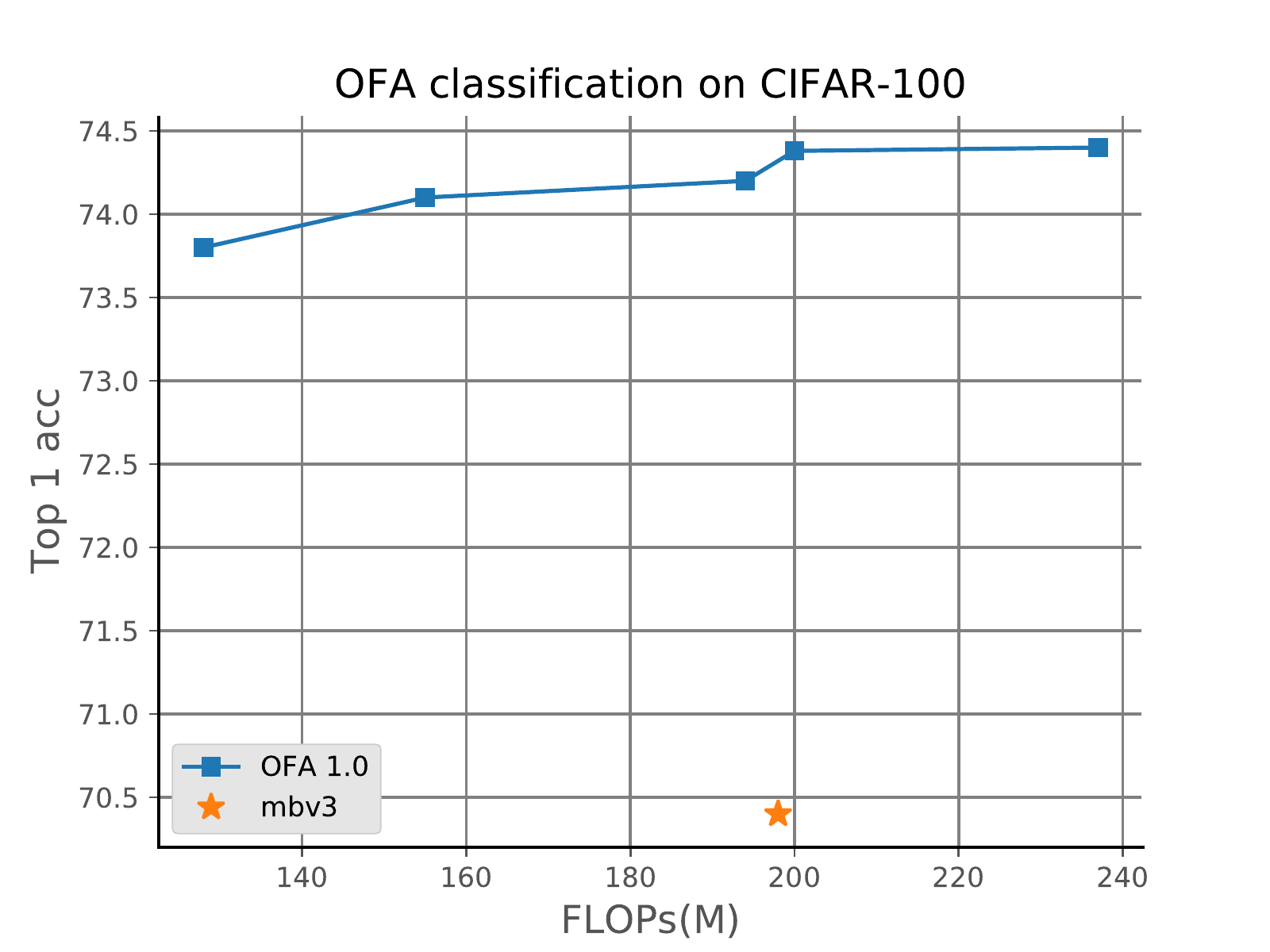}
    }
    \caption{OFA~\citep{cai2020once} classification results on CIFAR-10 and CIFAR-100. The search space is similar to that of MobileNet-V3~\citep{mobilenetv3}. Sub-networks are trained using Progressive Shrinking with Knowledge Distillation and finetuned after training. 0.8, 1.0, 1.2 in the legends denote the width multiplier.}
    \label{fig:ofa-classification}
    \vspace{-5pt}
\end{figure}

\begin{figure}[ht]
    \makebox[\textwidth][c]{
        \includegraphics[width=0.5\textwidth]{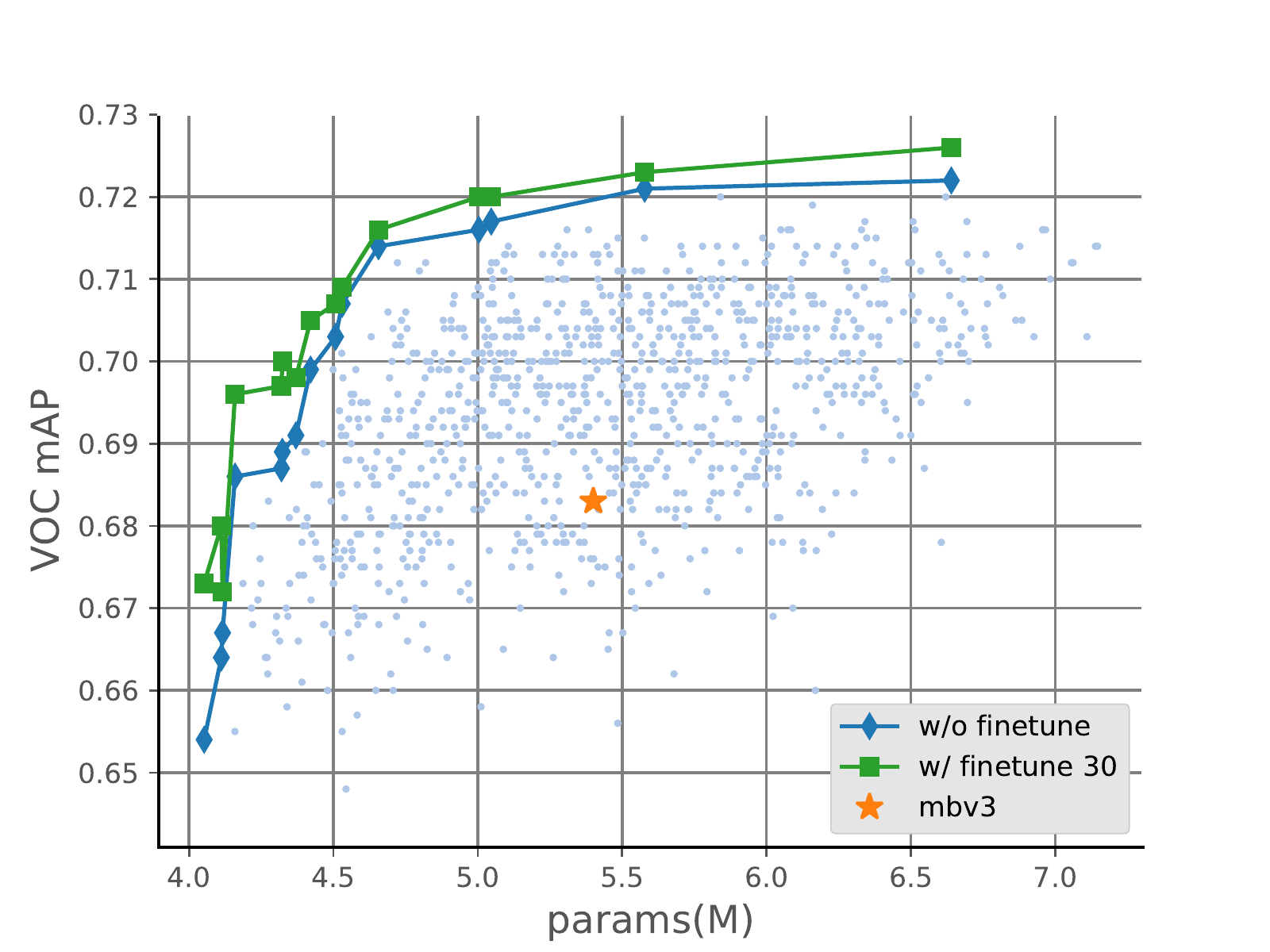}
    }
    \caption{OFA~\citep{cai2020once} detection result on VOC. The backbone search space is similar to that of MobileNet-V3~\citep{mobilenetv3}, and an SSD~\citep{liu2016ssd} head is used. Sub-networks are trained using Progressive Shrinking with Adaptive Distillation~\citep{tang2019learning} and (optionally) finetuned after training. In the search phase, 1k architectures are randomly sampled and tested (i.e., \textbf{random sample} controller is used).}
    \label{fig:ofa-detection}
\end{figure}

For more reproduction results, Fig.~\ref{fig:ofa-classification} shows the results of running OFA-based~\citep{cai2020once} search on CIFAR-10 and CIFAR-100. Due to the modularized design of \awnas, one can easily apply a methodology to new applications. Thus, based on the OFA methodology, we utilize \awnas to 
search for suitable backbones for object detection on the commonly-used VOC~\citep{voc} dataset, and show the results in Fig.~\ref{fig:ofa-detection}. The algorithm flow goes as 1) Supernet training phase: Train a supernet by calling ``awnas search'' without controller updates, in which the sub-networks using progressive shrinking with Adaptive Distillation. 2) Search phase: Identify the Pareto front by calling ``awnas search'' again without evaluator updates.

\begin{table}[ht]
  \centering
  \caption{Some \awnas reproduced results on CIFAR-10}
  \label{tab:reproduce}
  \begin{tabular}{c|c|c|c|c}
    \toprule
    {Method} & {Search Time} & {Performance} & {Params (M)} & {FLOPs (M)} \\
    \midrule
    ENAS~\citep{enas}      &  06h 17m & 97.30\% & 4.2 & 1303 \\
    DARTS~\citep{darts}    &  09h 05m & 97.11\% & 2.59 & 826 \\
    SNAS~\citep{snas}      &  08h 03m & 97.02\% & 3.18 & 1029 \\
    PC-DARTS~\citep{pcdarts} &  02h 57m & 97.43\% & 4.26 & 1343 \\
    \bottomrule
  \end{tabular}
\end{table}

Currently, our colleagues have been using \awnas to finish various researches: 1) NAS for robust and efficient NN system at edge~\citep{zeng2020blackbs,li2020fttnas}. 
2) Understanding and improving NAS algorithms~\citep{ning2020surgery,ning2020generic}.








\newpage
\section*{Appendix B. Hardware Profiling Pipeline and Cost Prediction Models}

Hardware-aware neural architecture search is critical for real-world tasks, especially for resource-constrained scenarios and real-time applications.
\awnas provides a set of tools and hardware cost models to support hardware-aware NAS. 
Namely, \awnas has a hardware profiling toolflow that enables primitive network generation, compilation, offline profiling, and result parsing. The profiling pipeline measures the latency and energy cost of search space primitives on CPU, GPU, and FPGA platforms.
From the profiled primitives' hardware cost, \awnas can accurately estimate the candidate network's latency and energy with a set of cost prediction models. Cost prediction models and hardware cost tables for CPU, GPU, and FPGA are released as hardware assets in \awnas.

\begin{figure}[ht]
    \begin{subfigure}{.32\textwidth}
    \centering
    \includegraphics[width=\linewidth]{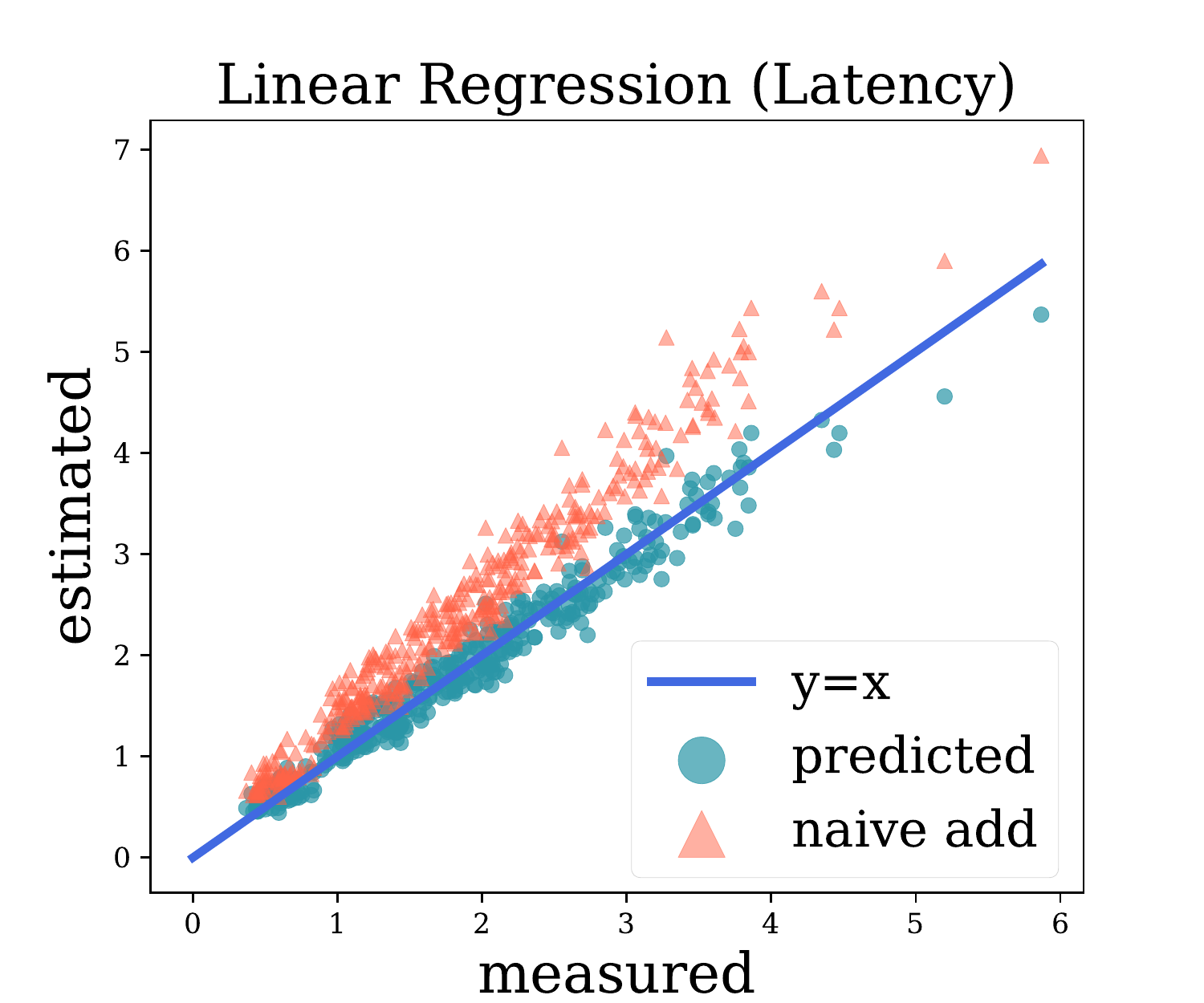}
    \caption{GPU latency, rMSE=0.174}
    \label{fig:latency-linear}
    \end{subfigure}
    \begin{subfigure}{.32\textwidth}
    \centering
    \includegraphics[width=\linewidth]{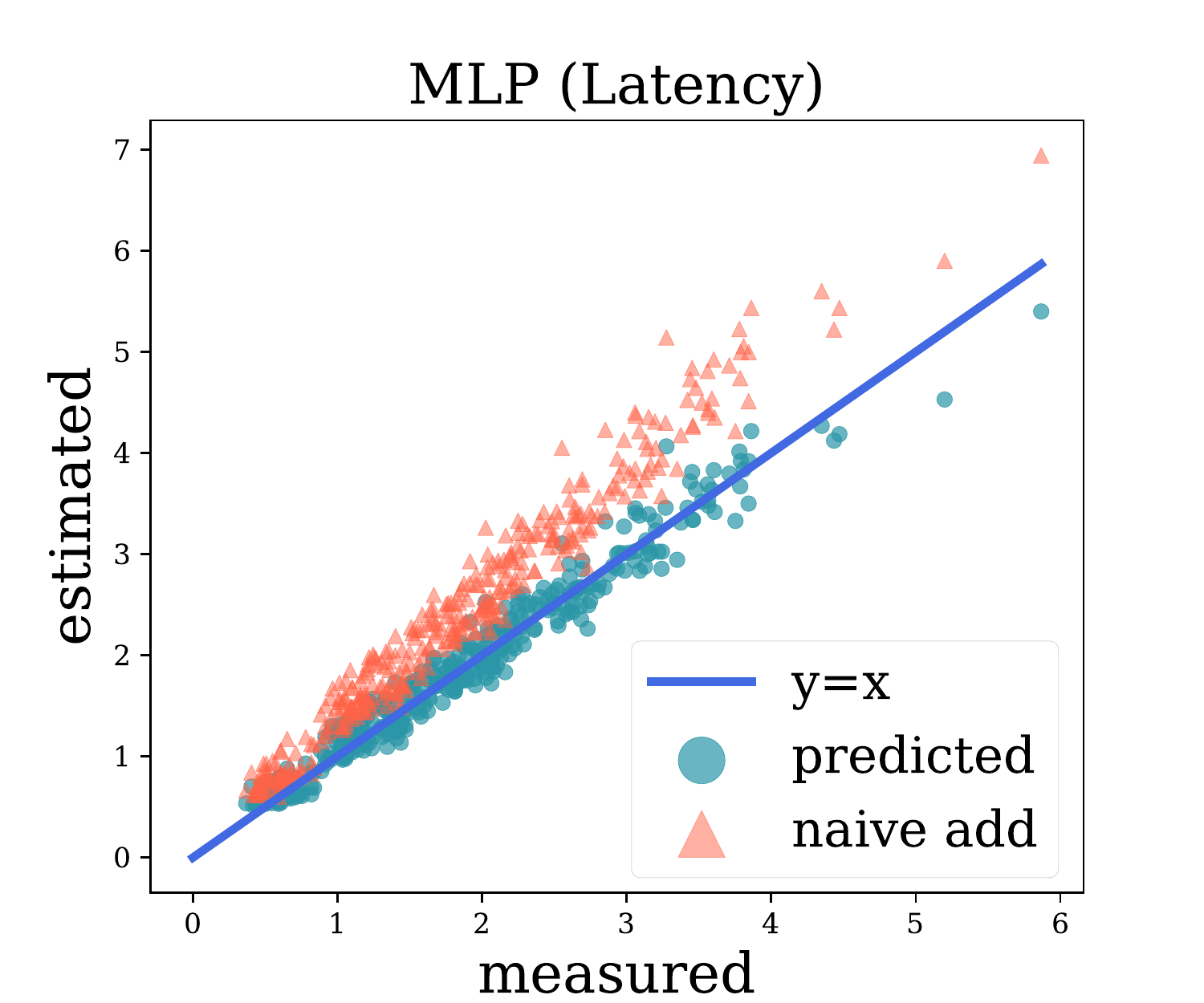}
    \caption{GPU latency, rMSE=0.174}
    \label{fig:latency-mlp}
    \end{subfigure}
    \begin{subfigure}{.32\textwidth}
    \centering
    \includegraphics[width=\linewidth]{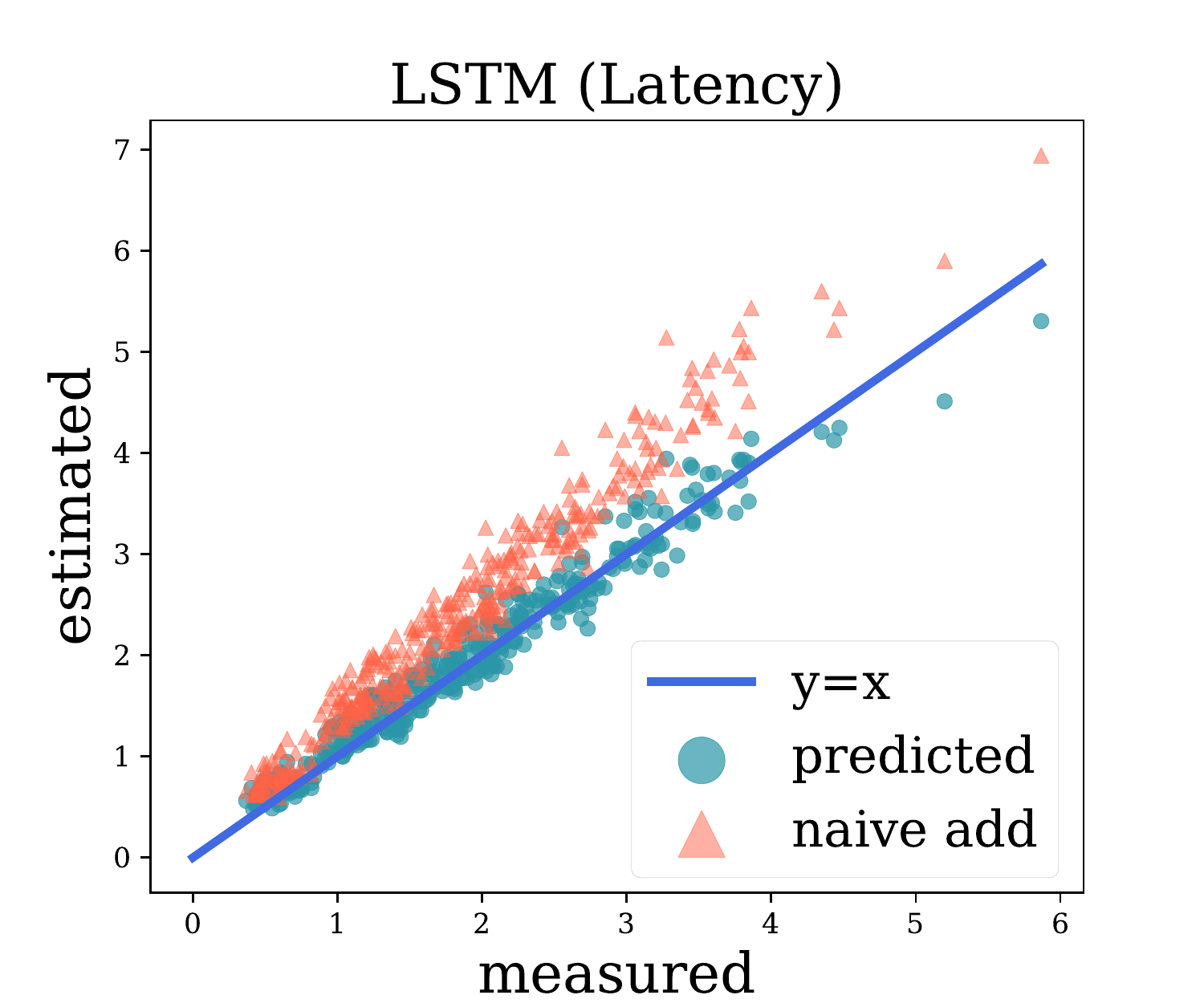}
    \caption{GPU latency, rMSE=0.187} 
    \label{fig:latency-lstm}
    \end{subfigure}
    \newline
    %
    \begin{subfigure}{.32\textwidth}
    \centering
    \includegraphics[width=\linewidth]{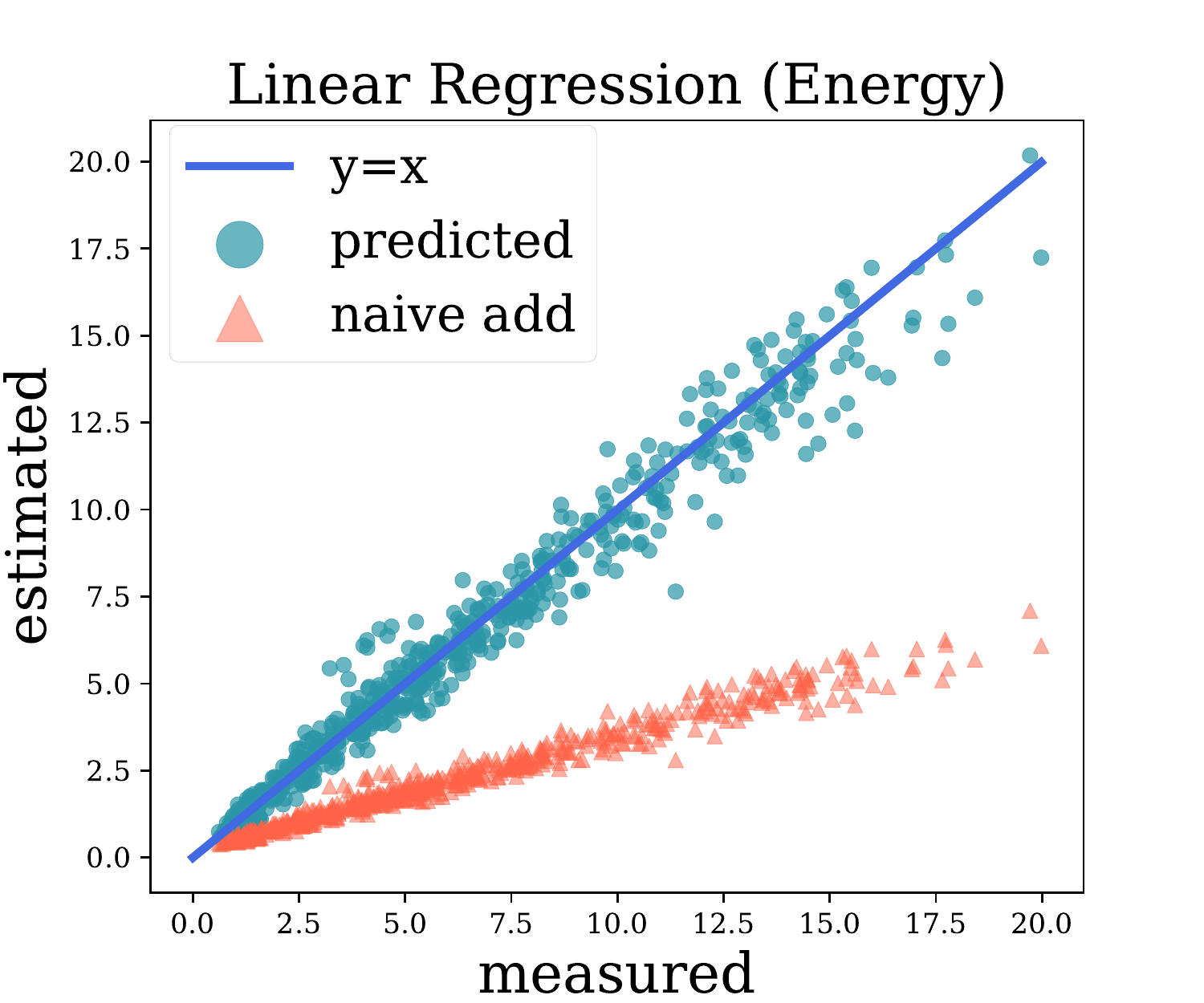}
    \caption{FPGA energy, rMSE=0.709}
    \label{fig:energy-linear}
    \end{subfigure}
    \begin{subfigure}{.32\textwidth}
    \centering
    \includegraphics[width=\linewidth]{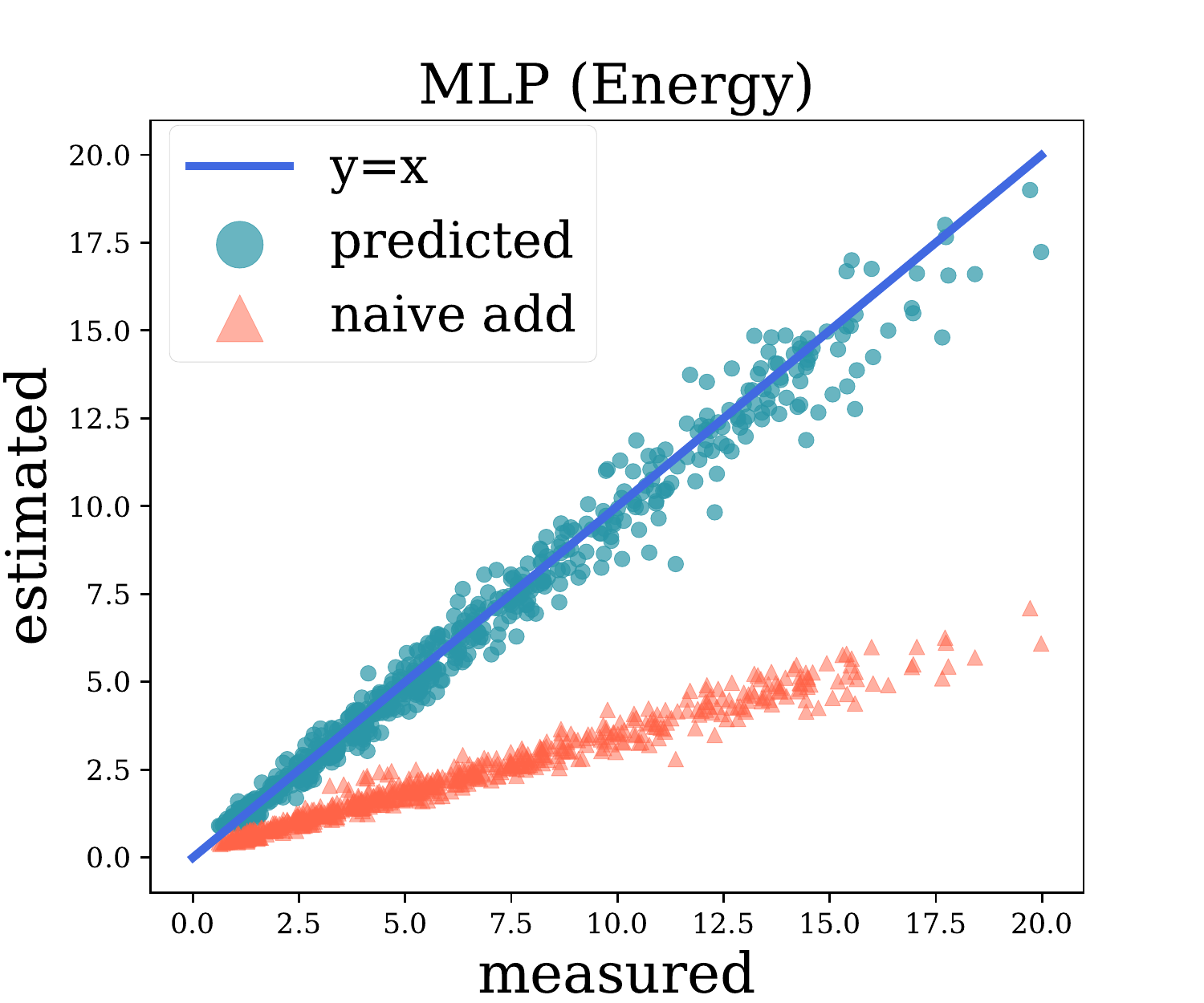}
    \caption{FPGA energy, rMSE=0.603}
    \label{fig:energy-mlp}
    \end{subfigure}
    \begin{subfigure}{.32\textwidth}
    \centering
    \includegraphics[width=\linewidth]{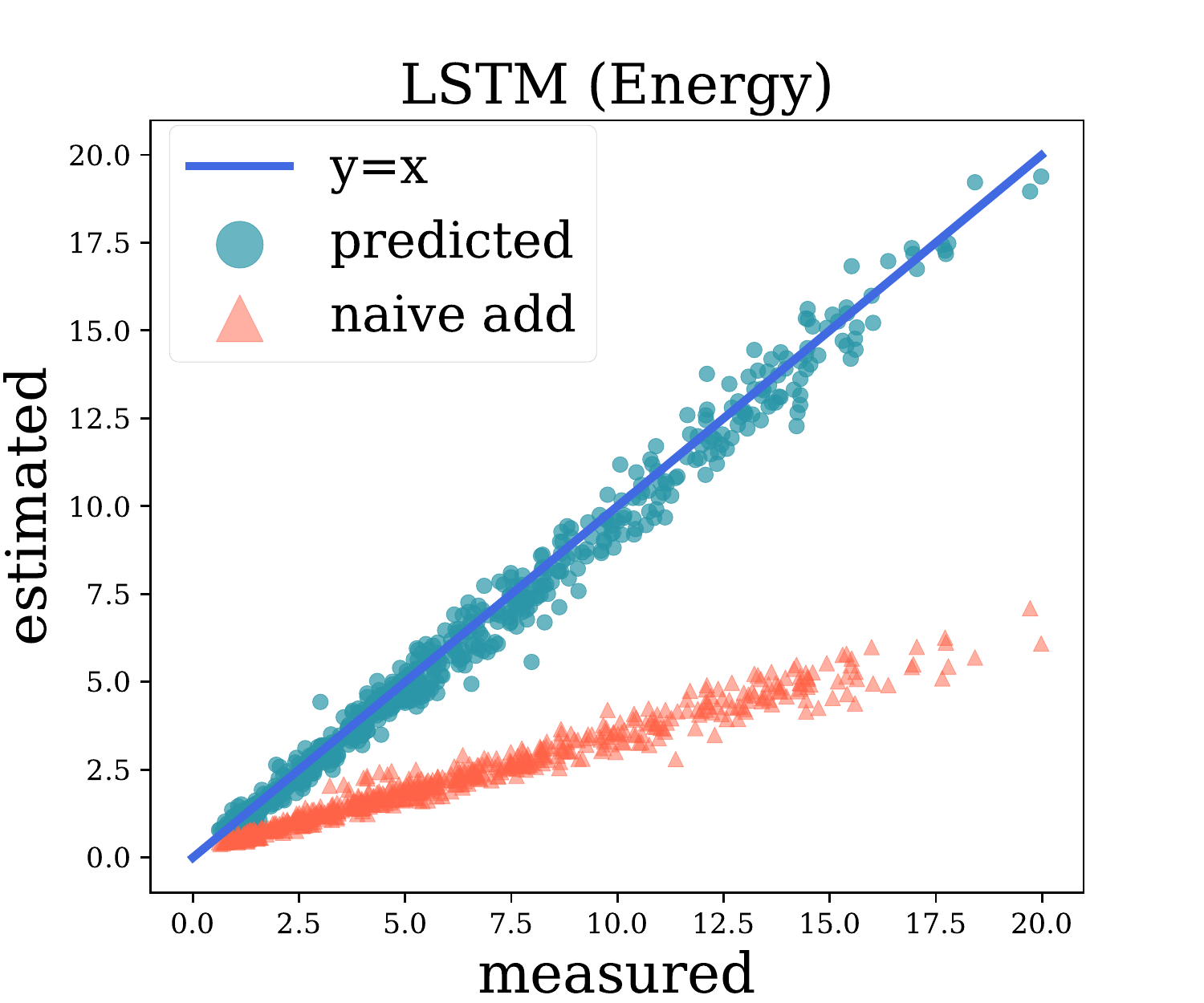}
    \caption{FPGA energy, rMSE=0.51} 
    \label{fig:energy-lstm}
    \end{subfigure}
    
\caption{GPU latency and FPGA energy estimation using three types of prediction models (Linear regression, MLP, LSTM). For GPU latency, estimation by naively adding up block latency results in rMSE=0.63. For FPGA energy, naive addition results in rMSE=4.94. Using the prediction models, we can achieve 3.6$\times$ and 9.7$\times$ better rMSE for GPU latency and FPGA energy respectively. 
}
\label{fig:hardware-model}
\end{figure}

\vspace{-0.2in}

Cost prediction models are necessary because deploying all the candidate networks in NAS to a target platform is often cost-prohibitive. Moreover, the primitives' latency and energy do not always add up to the candidate network's latency and energy. On devices such as FPGA where neural networks are executed sequentially, the sum of the building blocks' latency can approximate the overall network latency to a large extent. However, on platforms with complex cache mechanism and massive parallelism such as CPUs and GPUs, the summation of block latencies can significantly deviate from the actual network latency. Energy estimation on FPGA shares the same non-linear property because the network's power does not equal the summation of its building blocks' power.

\begin{figure}[th]
    \centering
    \includegraphics[width=0.8\textwidth]{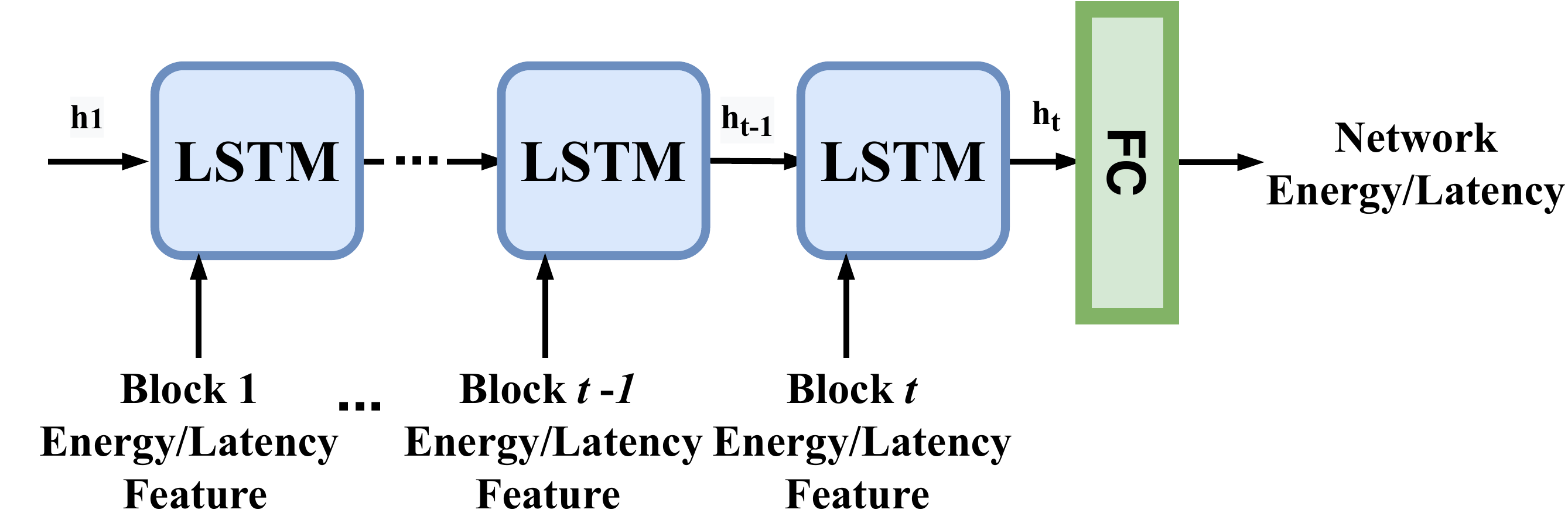}
    \caption{LSTM latency/energy hardware cost prediction model.}
    \label{fig:lstm}
    \vspace{-10pt}
\end{figure}

\begin{table*}[h]
\centering
\begin{tabular}{c|c|c|c}
\toprule
\textbf{Platform}  & \textbf{CPU}  & \textbf{GPU} & \textbf{FPGA} \\ \midrule
\textbf{Device} & Intel Xeon Gold 5115 & RTX-2080Ti &     Xilinx ZCU102               \\ \hline
\textbf{Metrics}   & Latency (ms)     & Latency (ms)  & \begin{tabular}[c]{@{}c@{}}Latency (ms)\\ Energy (mJ)\end{tabular} \\ \hline
\textbf{Profiling Tool}   & PyTorch  & PyTorch, CuDNN  & \begin{tabular}[c]{@{}c@{}} Xilinx Vitis \\ Power Advantage Tool \end{tabular} \\ \bottomrule
\end{tabular}
\caption{Hardware platform details for cost profiling}
\label{tab:hardware_details}
\end{table*}

Three types of prediction models are currently available in \awnas: linear regression model (1-variant or 2-variant), multilayer perceptron model (MLP), and LSTM-based model. For the single-variant linear model, the model takes the summation of primitives' latency/energy as input. For the 2-variant linear model, the input is the summation and block number. MLP model takes a vector of primitive latency as input and predicts the latency/energy for the candidate network. The structure of the LSTM model is illustrated in Fig.~\ref{fig:lstm}. At each time step, LSTM takes in a feature vector of block latency/energy and block configuration. More specifically, the block configuration consists of the input and output shape, kernel size, and stride. After all block features are processed, the final hidden state vector is fed into a fully-connected layer, which outputs the latency/energy prediction for the candidate network.

We conduct some experiments using \awnas's cost prediction models for estimating GPU latency and FPGA energy in the MobileNet-V2 search space (Fig.~\ref{fig:hardware-model}). The hardware platform details are presented in Table~\ref{tab:hardware_details}. We adopt the once-for-all~(\cite{cai2020once}) MobileNet-V2 search space design in the experiment. Specifically, a supernet with five stages is trained first, and candidate subnets are sampled from the supernet. Each stage in the supernet consists of numerous MobileNet-V2 inverted bottleneck blocks. When deriving candidate networks from the supernet, the number of blocks in each stage can be chosen from $\{2, 3, 4\}$ For each MobileNet-V2 block, the expansion ratio can be chosen from $\{3, 4, 6\}$, and the kernel size can be chosen from $\{3, 5, 7\}$.

Each model is trained with 2k random samples and tested on another 1k samples. Fig.~\ref{fig:latency-linear}-\ref{fig:latency-lstm} show the prediction versus ground truth of GPU latency on test dataset, and Fig.~\ref{fig:energy-linear}-\ref{fig:energy-lstm} show the prediction results for FPGA energy.
We observe a strong correlation between the estimated cost and the ground-truth. For GPU latency, the estimation rMSE ranges from 0.174 to 0.187, which is about 3.6$\times$ better than naive addition.
For FPGA energy estimation, prediction models achieve up to 9.7$\times$ improvement from naive addition.


\end{document}